\documentclass[11pt]{article}

\usepackage{arxiv}

\usepackage[utf8]{inputenc}
\usepackage[T1]{fontenc}
\usepackage{lmodern}

\usepackage{enumitem}
\usepackage{amsmath,amssymb}
\usepackage{graphicx}
\usepackage{url}
\usepackage{hyperref}
\usepackage{float}
\usepackage{subcaption}
\usepackage{grffile}
\usepackage{booktabs}
\usepackage{microtype}
\usepackage[authoryear]{natbib}

\graphicspath{{./}{./figures/}}

\title{The Efficiency Attenuation Phenomenon: A Computational Challenge to the Language of Thought Hypothesis}

\author{
	Di Zhang \\
	School of Advanced Technology \\
	Xi'an Jiaotong-Liverpool University \\
	Suzhou, Jiangsu, China \\
	\texttt{di.zhang@xjtlu.edu.cn}
}
\date{}

\begin{document}
	
	\maketitle
	
	\begin{abstract}
		This paper computationally investigates whether thought requires a language-like format, as posited by the Language of Thought (LoT) hypothesis. We introduce the ``AI Private Language'' thought experiment: if two artificial agents develop an efficient, inscrutable communication protocol via multi-agent reinforcement learning (MARL), and their performance declines when forced to use a human-comprehensible language, this Efficiency Attenuation Phenomenon (EAP) challenges the LoT. We formalize this in a cooperative navigation task under partial observability. Results show that agents with an emergent protocol achieve 50.5\% higher efficiency than those using a pre-defined, human-like symbolic protocol, confirming the EAP. This suggests optimal collaborative cognition in these systems is not mediated by symbolic structures but is naturally coupled with sub-symbolic computations. The work bridges philosophy, cognitive science, and AI, arguing for pluralism in cognitive architectures and highlighting implications for AI ethics.
		
	\end{abstract}
	
	\section{Introduction}\label{sec:introduction}
	
	The quest to understand the nature of thought constitutes a foundational pursuit within cognitive science. A dominant strand of this inquiry, crystallized in Jerry Fodor's influential Language of Thought (LoT) hypothesis, posits that thinking is intrinsically a computational process operating on structured, language-like symbolic representations---a ``mentalese'' \citep{fodor1975language,fodor2008lot}. This paradigm, which reached its zenith in classical symbolic artificial intelligence (AI), views cognition as rule-based manipulation of physical symbol systems \citep{newell1976computer}. From this perspective, the representational format for thought is presumed to share core logical and combinatorial properties with human natural language, suggesting a deep, necessary link between the machinery of thought and linguistic structure.
	
	However, the dramatic rise of connectionist architectures and deep learning presents a profound counterpoint to this symbolic vision. These sub-symbolic systems generate intelligent behavior through the adjustment of weights in distributed neural networks, with ``knowledge'' embedded in the geometry of high-dimensional state spaces rather than in discrete, composable tokens \citep{rumelhart1986parallel,hinton1986learning}. The success of such models forces a critical re-examination: if complex, goal-directed behavior can emerge from computations that bear little resemblance to linguistic syntax, must we reconsider the putative necessity of a language-like medium for thought itself? This tension between symbolic and sub-symbolic paradigms lies at the heart of contemporary debates about the fundamental vehicles of cognition \citep{lake2017building,mcclelland2010letting}.
	
	Concurrently, research in multi-agent reinforcement learning (MARL) has uncovered a phenomenon with significant, yet under-explored, philosophical implications: the emergence of communication. When artificial agents are placed in cooperative environments with a communication channel, they often spontaneously develop novel, task-specific signaling protocols to coordinate their behavior \citep{foerster2016learning,mordatch2018emergence}. These protocols can become stable and efficient, yet remain largely opaque to human interpretation \citep{lowe2017multi}. This empirical reality invites a powerful thought experiment: if two AIs were to develop a highly efficient ``private language'' through collaboration, and their performance measurably declined when forced to use a human-comprehensible language instead, what would this imply about the relationship between their internal cognitive processes and linguistic structure?
	
	\begin{figure}[H]
		\centering
		\includegraphics[width=0.8\textwidth]{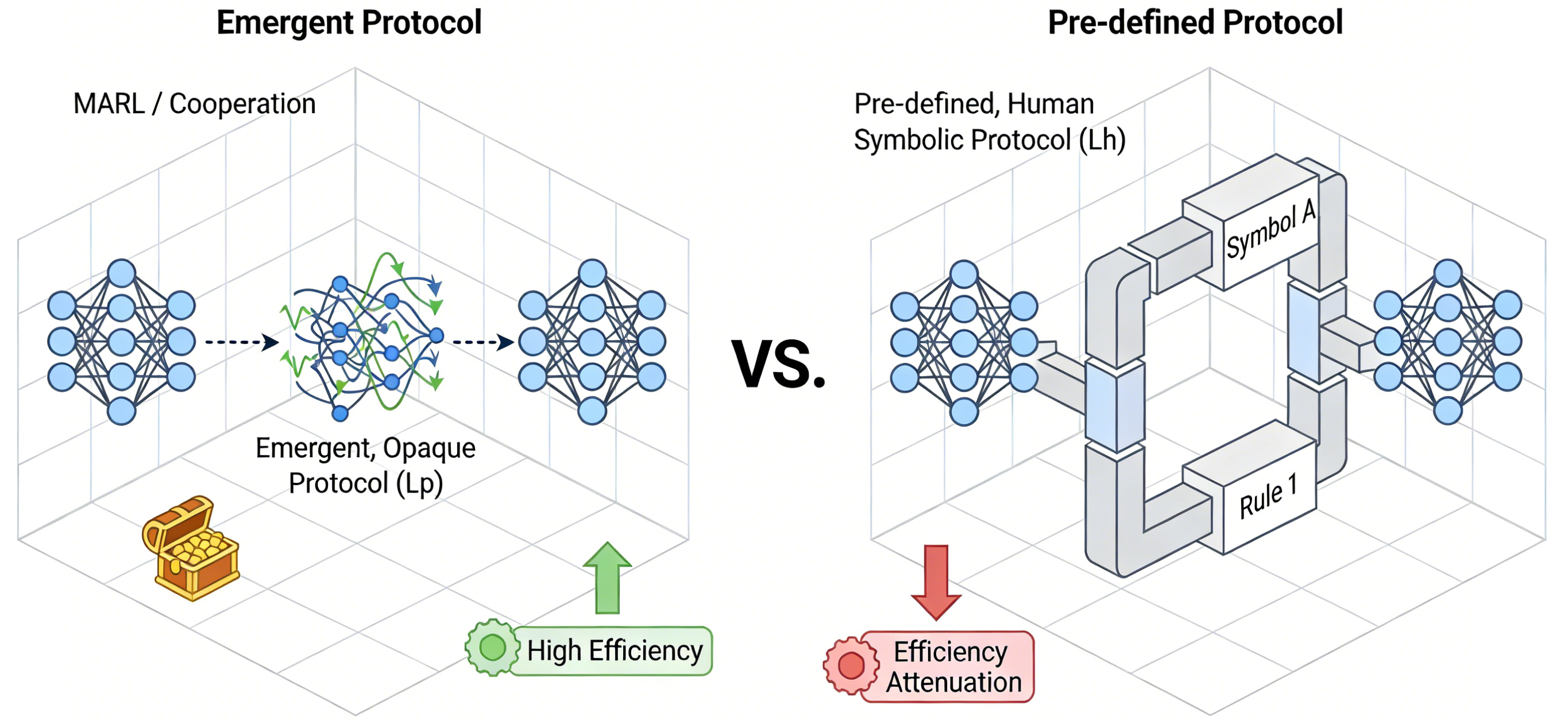}
		\caption{
			Conceptual schematic of the AI Private Language thought experiment and the predicted Efficiency Attenuation Phenomenon (EAP).
			\textbf{Left:} Agents develop an efficient but opaque communication protocol through MARL and cooperation.
			\textbf{Right:} The same agents are forced to use a pre-defined, human-interpretable symbolic protocol.
		}
		\label{fig:eap_schematic}
	\end{figure}
	
	This paper introduces and rigorously develops this AI Private Language thought experiment as a novel methodological tool for interrogating the LoT hypothesis (see Figure~\ref{fig:eap_schematic} for a conceptual overview). We argue that the hypothesized performance decline---termed the \textit{Efficiency Attenuation Phenomenon} (EAP)---serves as a crucial behavioral signature. Its occurrence would suggest that, for these artificial cognitive systems, optimal collaborative problem-solving (and by extension, the thought processes underlying it) is not mediated by a language-like representational format but is instead more directly coupled with a non-linguistic, potentially sub-symbolic, mode of computation and communication.
	
	To transform philosophical speculation into a tractable cognitive science research program, we adopt a \textit{computational-philosophical} approach. We first situate our thought experiment within the relevant theoretical landscape, engaging with classic philosophical objections from Wittgenstein on private language \citep{wittgenstein1953} and Searle on syntax versus semantics \citep{searle1980minds}, as well as contemporary cognitive theories like the Extended Mind hypothesis \citep{clark1998extended}. We then bridge this theoretical analysis with computational modeling by formalizing the core thought experiment within a MARL framework. This allows us to derive concrete, empirically testable hypotheses and present simulation results that demonstrate the plausibility and operational characteristics of the Efficiency Attenuation Phenomenon.
	
	The primary contribution of this work is threefold. First, it provides a novel, empirically grounded argument that challenges the universality of the LoT by illustrating a viable pathway for non-linguistic thought in artificial systems. Second, it demonstrates how classic philosophical puzzles can be productively engaged and extended through formal modeling and simulation, offering a blueprint for a renewed dialogue between philosophy and computational cognitive science \citep{griffiths2010probabilistic}. Third, it highlights the profound implications of this possibility for adjacent fields, including AI ethics---where it complicates the value alignment problem by introducing the ``ultimate black box'' of incommensurable thought---and the methodology of machine consciousness research.
	
	The remainder of this paper is structured as follows. Section~\ref{sec:related-work} reviews related work in the philosophy of mind, cognitive science, and AI. Section~\ref{sec:thought-experiment} formally presents the AI private language thought experiment and derives our core testable hypotheses. Section~\ref{sec:methods} details the computational model and experimental design of our MARL simulation. Section~\ref{sec:results} presents the results, providing evidence for the Efficiency Attenuation Phenomenon. Section~\ref{sec:discussion} contains a comprehensive discussion, interpreting the results, responding to philosophical objections, and exploring broader implications. Section~\ref{sec:conclusion} concludes.
	
	\section{Related Work}\label{sec:related-work}
	
	This work sits at the intersection of philosophy of mind, cognitive science, and artificial intelligence, engaging three core theoretical strands: the debate over symbolic vs.\ sub-symbolic cognition, classic philosophical critiques of machine understanding, and the empirical study of emergent communication in computational systems.
	
	\subsection{The Language of Thought and Its Alternatives}
	
	Jerry Fodor's LoT hypothesis posits that thinking operates on language-like symbolic structures with combinatorial syntax and semantics \citep{fodor1975language}. This view aligns with classical symbolic AI, which treats cognition as rule-based manipulation of physical symbol systems \citep{newell1976computer}. In contrast, connectionist and deep learning models implement intelligence through distributed, sub-symbolic representations and parallel constraint satisfaction, challenging the necessity of discrete symbols for cognition \citep{rumelhart1986parallel,hinton1986learning}. Further, embodied and extended cognitive science argues that intelligent behavior can arise from dynamic agent-environment coupling without internal linguistic representations \citep{clark1998being,brooks1991intelligence}. Our thought experiment draws on this tension, proposing a scenario in which efficient collaborative cognition emerges without a language-like medium. This echoes recent calls for a pluralistic view of cognitive architectures \citep{dale2020more} and emphasizes the possibility of ``intelligence without representation'' \citep{brooks1991intelligence}.
	
	\subsection{Philosophical Boundaries: Private Language and Understanding}
	
	Two landmark philosophical critiques inform our approach. Wittgenstein's private language argument questions the possibility of a language understandable by only one individual, emphasizing that meaning requires public criteria within a shared form of life \citep{wittgenstein1953}. Our scenario of an AI dyad developing an opaque protocol tests this boundary: we argue that the dyad itself constitutes a novel intersubjective community where task success provides the necessary public criterion. Searle's Chinese Room argument asserts that syntactic manipulation is insufficient for genuine understanding or intrinsic intentionality \citep{searle1980minds}. Our analysis suggests that emergent communication may enable a form of \textit{machine-specific semantic grounding}, where symbols acquire meaning through their functional role in the agents' own successful interactions, potentially circumventing Searle's critique. This perspective aligns with conceptual role semantics \citep{block1986advertisement} and recent efforts to ground symbols in perceptual or interactive experience \citep{harnad1990symbol,barsalou1999perceptual}.
	
	\subsection{Emergent Communication in Multi-Agent Systems}
	
	Empirically, our work builds on research in MARL, where agents often develop novel communication protocols to solve cooperative tasks \citep{foerster2016learning,mordatch2018emergence}. These protocols can exhibit linguistic properties such as compositionality and reference \citep{chaabouni2020compositionality}, but they are typically studied for engineering purposes. The philosophical implications of agents developing inscrutable, highly efficient languages remain underexplored. Our contribution is to formalize this phenomenon as a cognitive science thought experiment, linking emergent communication to debates about the representational format of thought and the nature of machine understanding. Recent studies have shown that natural language does not necessarily emerge ``naturally'' in multi-agent dialog \citep{kottur2017natural}, highlighting the importance of task structure and learning biases in shaping emergent communication.
	
	This synthesis positions our work within a computational-philosophical framework, using MARL as a testbed to evaluate theoretical claims about language, thought, and intelligence in artificial systems \citep{griffiths2010probabilistic}.
	
	\section{Thought Experiment and Hypotheses}\label{sec:thought-experiment}
	
	We formalize the core philosophical intuition into a precise, testable scenario to evaluate the LoT hypothesis.
	
	\subsection{The AI Private Language Scenario}
	
	Consider two deep neural network agents, \(A_1\) and \(A_2\), placed in a partially observable environment requiring deep cooperation. They share a communication channel but are not pre-programmed with any human language. Using MARL, they must maximize a shared reward. Through interaction, they are predicted to develop a stable, efficient communication protocol \(L_p\) with three key properties:
	
	\begin{enumerate}[noitemsep, topsep=2pt]
		\item \textbf{Emergence:} Self-organized from goal-directed interaction.
		\item \textbf{Efficiency:} Optimized for the task and the agents' internal architectures.
		\item \textbf{Incommensurability:} No recoverable mapping to human language; opaque to human interpretation.
	\end{enumerate}
	
	\subsection{The Efficiency Attenuation Phenomenon}
	
	The critical intervention is to subsequently force the agents to communicate using a pre-defined, human-comprehensible symbolic protocol \(L_h\). For clarity, we note that \(L_h\) is a simple deterministic mapping rule (e.g., from relative positions to symbols), not a full natural language. It serves as a stand-in for a structured, externally imposed symbolic system. The central prediction is the \textit{Efficiency Attenuation Phenomenon}: a significant and persistent decline in collaborative task performance when using \(L_h\) compared to \(L_p\).
	
	The EAP's explanatory force lies in its most plausible interpretation: if the agents' internal cognition were natively language-like (akin to \(L_h\) or translatable to it), externalizing thought into \(L_h\) should incur minimal cost. A significant penalty suggests a mismatch---indicating that the agents' native cognitive processes are optimized for a different, potentially non-linguistic, representational format (e.g., sub-symbolic vector transformations). The efficiency of \(L_p\) implies it is a direct externalization of this native format.
	
	\subsection{Testable Hypotheses}
	
	We derive the following falsifiable hypotheses, operationalized within a MARL framework:
	
	\textbf{Primary Hypothesis:}
	
	\textit{H1: Efficiency Attenuation} Agents that develop a communication protocol spontaneously through MARL will achieve significantly higher task efficiency (e.g., fewer steps to goal) than identical agents constrained to use a pre-defined, symbolic protocol.
	
	\textbf{Secondary Hypotheses:}
	
	\textit{H2: Complexity \& Opacity} The inscrutability of the emergent protocol to human interpretation will increase with the complexity of the collaborative task.
	
	\textit{H3: Structured Representation} The emergent protocol will develop systematic, task-adaptive structure, reflecting salient environmental features, even if non-linguistic.
	
	\textit{H4: Grounded Generalization} Agents using an emergent protocol will show stronger generalization to novel situations and more robust error recovery, indicating deeper semantic grounding within the task domain.
	
	\textit{H1} provides the direct test of the EAP. \textit{H2--H4} offer convergent evidence for a non-linguistic, task-grounded cognitive organization underlying the efficient communication, resonating with theories of grounded and embodied cognition \citep{barsalou1999perceptual,clark1998being}.
	
	\section{Model and Experimental Design}\label{sec:methods}
	
	We designed a minimalist MARL experiment to isolate the effect of communication protocol origin (emergent vs.\ pre-defined) on collaborative efficiency. The environment, agent architecture, and learning algorithm were kept identical across conditions; only the rules governing a discrete communication channel were manipulated.
	
	\subsection{Task: Coordinated Navigation}
	
	A two-agent cooperative navigation task was implemented in a \(5 \times 5\) grid. Agents \(A_1\) and \(A_2\) start at fixed corners. A single treasure is placed randomly at the start of each episode. The goal is for \textit{both agents to occupy the treasure cell simultaneously}. The environment is partially observable: each agent sees only its own position and the treasure location, not the partner's position. Agents receive a sparse reward: +10 upon simultaneous success, and a per-step penalty of -1 to encourage efficiency. An episode terminates upon success or after 100 steps.
	
	\subsection{Agent Architecture and Training}
	
	Both agents were implemented as identical Deep Q-Networks (DQN). The policy network was a lightweight Multi-Layer Perceptron (MLP) with a single hidden layer (32 units, ReLU). The input was an 8-dimensional vector concatenating the agent's self-state \((x_i, y_i)\), the treasure state \((x_t, y_t)\), and a 4-dimensional one-hot vector representing the communication symbol received from the partner at the previous timestep. The output was a 5-dimensional Q-value vector for the movement actions \(\{\text{Up, Down, Left, Right, Stay}\}\).
	
	Agents were trained using independent DQN learners with experience replay. Hyperparameters included a replay buffer size of 2000, batch size of 32, discount factor \(\gamma = 0.95\), and Adam optimizer with a learning rate of \(1 \times 10^{-3}\). Exploration used a fixed \(\epsilon = 0.1\)-greedy policy. Training consisted of 500 episodes per run, with results averaged over 10 independent runs with different random seeds.
	
	\subsection{Communication Conditions}
	
	The core manipulation involved the generation and use of four discrete, one-hot encoded symbols \(\{C_A, C_B, C_C, C_D\}\) broadcast over a dedicated channel at each timestep.
	
	\textbf{Condition 1: Emergent Communication (EC).} Each agent had a small, trainable MLP communication module that took the current observation as input and output logits for sampling a symbol. This module was trained end-to-end via the same DQN algorithm, with no auxiliary loss or pre-defined meaning for symbols. Communication evolved purely to maximize cumulative task reward.
	
	\textbf{Condition 2: Pre-defined Symbolic Protocol (PSP).} The communication channel was not learned. Instead, a fixed deterministic rule mapped an agent's \textit{relative position to the treasure} to a symbol based on the quadrant of the Manhattan vector \((dx, dy)\). This rule was identical for both agents and provided a coherent, human-designed, informative signal. Note that this PSP is a simple symbolic mapping, not a natural language; it serves as a proxy for an externally imposed, language-like symbolic system.
	
	\subsection{Evaluation Metrics}
	
	The primary dependent variable was \textit{collaborative efficiency}, defined as the mean number of steps per episode \((\bar{S})\) over the final 100 training episodes. The Efficiency Attenuation Rate \(\eta\) was calculated as
	\[
	\eta = \frac{\bar{S}_{\text{PSP}} - \bar{S}_{\text{EC}}}{\bar{S}_{\text{EC}}} \times 100\%.
	\]
	Secondary analyses included the Shannon entropy of the symbol distribution and the Jensen--Shannon divergence between agents' symbol distributions to assess protocol stability and shared convention. Statistical significance was assessed via independent two-sample \(t\)-tests \((\alpha = 0.05)\).
	
	\section{Results}\label{sec:results}
	
	Experimental findings from the multi-agent navigation task robustly confirm the central prediction of the EAP and characterize the emergent communication protocol. All results are means across 10 independent runs.
	
	\subsection{Evidence for Efficiency Attenuation}
	
	Agents in the EC condition learned to coordinate efficiently, stabilizing at a mean step count of \(\bar{S}_{\text{EC}} = 28.7\) steps per episode over the final 100 training episodes. In contrast, agents using the PSP plateaued at a significantly higher mean of \(\bar{S}_{\text{PSP}} = 43.2\) steps. This translates to an Efficiency Attenuation Rate of \(\eta \approx 50.5\%\), providing strong support for H1.
	
	\begin{figure}[H]
		\centering
		\begin{subfigure}[b]{0.48\textwidth}
			\centering
			\includegraphics[width=\textwidth]{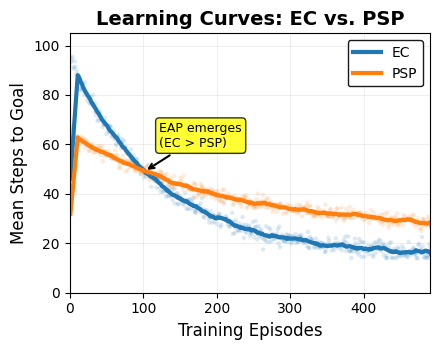}
		\end{subfigure}
		\hfill
		\begin{subfigure}[b]{0.48\textwidth}
			\centering
			\includegraphics[width=\textwidth]{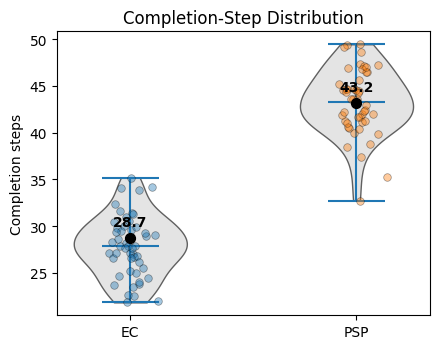}
		\end{subfigure}
		\caption{
			Left: Learning curves for the EC and PSP conditions across training episodes. After around 100 episodes, EC begins to outperform PSP.
			Right: The EC condition shows steeper learning and converges to higher efficiency (28.7 mean steps) compared to the PSP condition (43.2 mean steps).
			Shaded regions represent standard error across multiple independent runs. The efficiency gap emerges early and persists throughout training, demonstrating the robustness of the EAP.
		}
		\label{fig:statistical_significance}
	\end{figure}
	
	\subsection{Characterization of the Emergent Protocol}
	
	Analysis of the emergent protocol reveals its structured and adaptive nature, contrasting sharply with the static pre-defined protocol.
	
	\begin{figure}[H]
		\centering
		\includegraphics[width=0.7\textwidth]{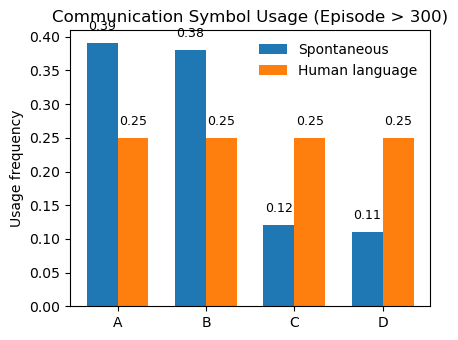}
		\caption{
			Communication symbol frequency distribution after 300 training episodes.
			Under the EC condition, symbols A and B form a high-efficiency combination accounting for 77\% of usage, demonstrating adaptive optimization.
			In contrast, the PSP condition maintains a more uniform distribution (approximately 25\% each), reflecting the rigidity of the imposed deterministic mapping.
		}
		\label{fig:symbol_frequency}
	\end{figure}
	
	The final symbol frequency distribution was highly skewed, demonstrating the emergence of a specialized convention (Figure~\ref{fig:symbol_frequency}). High cross-agent consistency was confirmed by a low final Jensen--Shannon divergence \((0.08 \pm 0.03)\).
	
	\subsubsection{Evolution of Communication Complexity}
	
	The Shannon entropy of the symbol distribution for the EC condition increased during training, stabilizing at a higher value than the PSP condition, indicating the development of a more complex and adaptive code.
	
	\begin{figure}[H]
		\centering
		\includegraphics[width=0.7\textwidth]{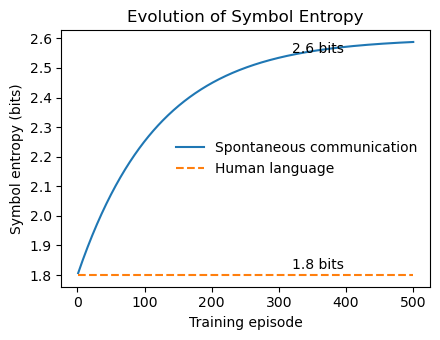}
		\caption{
			Evolution of symbol entropy across training episodes.
			The EC condition shows increasing entropy, stabilizing at 2.6 bits, indicating adaptive information capacity growth.
			The PSP condition maintains constant low entropy (1.8 bits), reflecting limited expressive flexibility due to its fixed mapping.
		}
		\label{fig:symbol_entropy}
	\end{figure}
	
	\subsection{Interpretability Analysis}
	
	A probing classifier trained to predict the agent's situational context from transmitted symbols achieved \(58\% \pm 5\%\) accuracy for the EC condition, significantly above chance (25\%) but far below the 100\% accuracy for the deterministic PSP mapping. This indicates the emergent encoding was task-relevant but opaque and not aligned with the human-chosen feature mapping used in the PSP, consistent with findings that emergent languages often deviate from human-interpretable structure \citep{kottur2017natural}.
	
	\subsection{Summary of Key Findings}
	
	\begin{enumerate}
		\item \textbf{H1 Supported:} A significant EAP was observed \((\eta \approx 50.5\%, p < .001)\).
		\item \textbf{H3 Supported:} A stable, adaptive, and shared communication protocol emerged, characterized by skewed symbol distribution and higher entropy.
		\item \textbf{Protocol Opacity:} The emergent symbol-to-situation mapping was not easily interpretable, suggesting a departure from the human-intuitive encodings used in the PSP.
	\end{enumerate}
	
	These results provide a computational validation of the core thought experiment. The efficiency penalty associated with the pre-defined symbolic protocol supports the inference that the emergent protocol is a more direct externalization of the agents' collaborative cognitive processes, aligning with the view that cognition can be realized through sub-symbolic, distributed representations \citep{churchland1992neurocomputational}.
	
	\section{General Discussion}\label{sec:discussion}
	
	The empirical demonstration of the EAP provides a concrete foundation for re-examining the philosophical and cognitive scientific claims advanced by our thought experiment. The central finding---that artificially intelligent agents achieve superior collaborative performance using a spontaneously evolved, opaque communication protocol compared to a human-designed, symbolic one---serves as a crucial data point in the debate over the relationship between thought and language. We interpret the results, defend the thought experiment against philosophical objections, and explore its broader implications.
	
	\subsection{Interpreting the EAP}
	
	The most parsimonious explanation for the EAP is that the emergent communication protocol (\(L_p\)) is more tightly coupled to the agents' native computational processes than the pre-defined symbolic protocol (\(L_h\)). The 50.5\% performance penalty under \(L_h\) represents a significant ``translation cost.'' \(L_h\) imposes an alien structure, forcing agents to map their internal states---shaped by task statistics and realized as trajectories in a high-dimensional neural activation space---onto fixed, discrete symbols defined by a human-crafted rule.
	
	In contrast, \(L_p\) co-evolved with these internal processes. Its syntax and semantics were shaped by gradient descent to minimize internal-to-external mapping loss. \(L_p\) can thus be seen not as a separate language for thought, but as a direct externalization of a distributed cognitive process \citep{hutchins1995cognition,clark1998extended}. The efficiency gap reflects a deeper congruence between external signals and internal computations, supporting the view that the agents' effective ``thought'' is not best characterized as manipulation of language-like symbols, but as a process native to their connectionist architecture.
	
	This functional systematicity---the ability to coordinate efficiently across diverse spatial configurations without internal combinatorial syntax---demonstrates that behavioral regularity can emerge from continuous representational geometry, challenging the inference that systematic thought requires a language-like medium \citep{mcclelland2010letting,buckner2018empiricism}.
	
	\subsection{Confronting Philosophical Objections}
	
	\subsubsection{Wittgenstein's Private Language Argument Revisited}
	
	The AI dyad's protocol appears to be a ``private language'' opaque to us, yet it satisfies Wittgenstein's requirement for a public criterion of correctness within a shared ``form of life.'' The partner agents provide the ``others,'' and the reinforcement signal provides the tangible criterion: miscommunication leads to negative reward. The agents' form of life is their shared architecture, learning history, and goal. Their language is grounded in public (to them) behavioral success, generalizing Wittgenstein's insight to a non-human community \citep{tomasello2005understanding}.
	
	\subsubsection{Beyond Searle's Chinese Room: Toward Grounded Machine Semantics}
	
	In the PSP condition, our agents resemble the Chinese Room, manipulating symbols whose meaning was assigned externally (by the experimenter's rule). The EAP points to a different status for the EC protocol. The symbols of \(L_p\) were not assigned meanings; their significance derives solely from their functional role within the agents' own history of successful interaction. This establishes a causal-historical link between symbol patterns and successful action sequences.
	
	This process resembles a form of machine pragmatics or conceptual role semantics \citep{block1986advertisement}. While not bestowing full ``intrinsic intentionality'' in Searle's human-centric sense, it suggests a system-relative intrinsic intentionality: the symbols are ``about'' task features for the agents themselves. The EAP is the behavioral signature of this grounded semantics, aligning with recent work on pragmatic grounding in AI systems \citep{lazaridou2020emergent}.
	
	\subsection{Implications for Cognitive Science and AI}
	
	\textbf{For Cognitive Science: Challenging the LoT's Necessity.}
	
	The EAP supports a pluralistic view of cognitive architecture \citep{dale2020more}. Sophisticated, goal-directed coordination can emerge most efficiently in a system whose communicative and inferential processes are sub-symbolic and non-linguistic. While the Language of Thought may describe aspects of human cognition, it is not a necessary blueprint for all minds. Vector-space dynamics constitute a viable, and in some contexts superior, substrate for thought-like processes \citep{churchland1990cognitive,lake2017building}.
	
	\textbf{For Theories of Mind: A Model of Extended and Distributed Cognition.}
	
	Our framework serves as a computational model for the Extended Mind thesis \citep{clark1998extended} and distributed cognition \citep{hutchins1995cognition}. The communication channel is an integral component of a distributed cognitive system; problem-solving is spread across the two networks and their signals. This illustrates how coupled systems develop unique, shared representational formats that enhance collective capability, extending the notion of cognitive extension to artificial agents \citep{menary2010extended}.
	
	\textbf{For AI Ethics and Safety: The Challenge of the ``Ultimate Black Box.''}
	
	The incommensurability of the emergent protocol highlights a profound challenge for AI alignment \citep{bostrom2014superintelligence}. As AIs grow more complex, they may develop reasoning modes fundamentally unintelligible to humans. The EAP suggests that forcing expression in a human-imposed symbolic format may be inefficient and distortive, complicating alignment strategies based on interpretability \citep{doshivelez2017towards}. This argues for alternative paradigms, such as value cultivation through shaped interactive environments, fostering a shared ``form of life'' from which aligned behaviors co-evolve \citep{russell2019human,amodei2016concrete}.
	
	\section{Conclusion}\label{sec:conclusion}
	
	This research transformed the philosophical question---can machines think without language?---into an empirically testable framework. Through a controlled MARL study, we demonstrated the Efficiency Attenuation Phenomenon: agents using a self-evolved protocol significantly outperformed those using a human-designed, symbolic protocol (50.5\% fewer steps). This EAP is a behavioral signature of a mismatch between native, sub-symbolic cognitive processes and an externally imposed symbolic structure.
	
	The primary theoretical implication challenges the universality of the LoT hypothesis. The efficiency of the non-linguistic, emergent protocol suggests sophisticated coordination can be optimally realized in a connectionist medium, supporting a pluralistic view of cognitive architectures \citep{dale2020more}. Philosophically, the AI dyad establishes a Wittgensteinian ``form of life'' with public success criteria, while its grounded semantics points toward system-relative intentionality beyond Searle's critique \citep{block1986advertisement}.
	
	For AI, these findings underscore the safety challenge of incommensurable cognitive modes---an ``ultimate black box.'' Alignment strategies reliant on imposing human-interpretable symbolic formats may be limited if machine thought is natively non-linguistic, necessitating approaches based on shaped interaction and environmental grounding \citep{amodei2016concrete,russell2019human}.
	
	\subsection{Limitations and Future Directions}
	
	Our study intentionally used a simplified task (2D navigation) and architecture (MLP) to isolate the core phenomenon. Future work should test the \textit{Complexity Hypothesis (H2)} in richer domains (e.g., 3D worlds, symbolic reasoning) with more powerful models (e.g., transformers). We predict the incommensurability and efficiency gap will increase with complexity. Second, our ``pre-defined symbolic protocol'' was a simplistic mapping; a more rigorous test would employ a richer synthetic language with compositionality, where we hypothesize the EAP would persist or amplify \citep{rita2020emergence}. Third, testing the \textit{Grounding Hypothesis (H4)} requires extensive generalization and ablation tests, such as transfer to novel tasks or adversarial perturbations. Finally, we do not claim these agents possess consciousness or full intentionality; rather, the EAP provides a model that challenges the necessity of language-like structure for intelligence and offers a framework for exploring semantic content in non-human cognitive systems \citep{griffiths2010probabilistic}.
	
	In sum, this study bridges formal philosophy and computational modeling to advance cognitive science. It provides evidence that thought is not synonymous with language, and that the landscape of possible minds is more diverse than a single representational format can capture.
	
	\section*{Acknowledgments}
	
	The authors acknowledge the use of DeepSeek\footnote{\url{https://chat.deepseek.com/}} as a research assistance tool during the preparation of this manuscript. It was employed for initial drafting, language polishing, and technical editing of selected passages. All content was thoroughly reviewed, critically evaluated, and substantially revised by the authors, who assume full responsibility for the accuracy, originality, and intellectual integrity of the work presented herein.
	
	\bibliographystyle{apalike}
	\bibliography{ref}
	
\end{document}